\documentclass[runningheads,a4paper]{llncs}

\usepackage{amssymb}
\usepackage{graphicx}

\usepackage{url}
\urldef{\mails}\path|{sai.qian, maxime.amblard}@loria.fr|
\newcommand{\keywords}[1]{\par\addvspace\baselineskip
\noindent\keywordname\enspace\ignorespaces#1}

\usepackage{stmaryrd}
\usepackage{multirow}
\usepackage{linguex}

\usepackage{enumerate}
\title{Event in Compositional Dynamic Semantics}

\author{Sai Qian\inst{1,2,3}, Maxime Amblard\inst{1,2} }

\institute{
LORIA \& INRIA Nancy Grand-Est  - BP 239 - 54506 Vand\oe uvre-l\`es-Nancy Cedex 
\and Nancy 2 University, 13 Rue du Maréchal Ney - 54037 Nancy Cedex 
\and Nancy 1 University, 24-30 Rue Lionnois - BP 60120 $|$ 54003 Nancy Cedex
\\ \mails
}

\begin{document}

%\mainmatter  % start of an individual contribution


% a short form should be given in case it is too long for the Cryning head
%\titleCryning{Lecture Notes in Computer Science: Authors' Instructions}

% the Cryning heads and the author index.
%

%
%\authorCryning{Lecture Notes in Computer Science: Authors' Instructions}

%%%\institute{Nancy 1 University\\
%Tiergartenstr. 17, 69121 Heidelberg, Germany\\
%\url{http://www.springer.com/lncs}
%%%}


%\toctitle{Lecture Notes in Computer Science}
%\tocauthor{Authors' Instructions}
\maketitle

\begin{abstract}
We present a framework which constructs an event-style discourse semantics. The discourse dynamics are encoded in continuation semantics and various rhetorical relations are embedded in the resulting interpretation of the framework. We assume discourse and sentence are distinct semantic objects, that play different roles in meaning evaluation. Moreover, two sets of composition functions, for handling different discourse relations, are introduced. The paper first gives the necessary background and motivation for event and dynamic semantics, then the framework with detailed examples will be introduced.
%The structure of this paper goes on as following: first we give the background and motivation for using event-based semantics. Then we talk about the transition from static to dynamic semantics. Two dynamic formalisms, the continuation semantics in \cite{philippe} and SDRT will be presented. Finally we introduce a semantic framework which combines event, dynamics and discourse structure.
%The abstract should summarize the contents of the paper and should contain at least 70 and at most 150 words. It should be written using the \emph{abstract} environment.
%\keywords{We would like to encourage you to list your keywords within the abstract section}
\end{abstract}

\section{Event Semantics}\label{event_intro}

%\subsection{Brief History}

The idea of relating verbs to certain events or states can be found throughout the history of philosophy. For example, a simple sentence \textit{John cries} can be referred to a $crying$ action, in which $John$ is the agent who carries out the action. However, there were no real theoretical foundations for semantic proposals based on events before \cite{davidson67lf}. In \cite{davidson67lf}, Davidson explained that a certain class of verbs (action verbs) explicitly imply the existence of underlying events, thus there should be an implicit event argument in the linguistic realization of verbs.

For instance, traditional Montague Semantics provides \textit{John cries} the following interpretation: $Cry(john)$, where $Cry$ stands for a 1-place predicate denoting the $crying$ event, and $john$ stands for the individual constant in the model. Davidson's theory assigns the same sentence another interpretation: $\exists e.Cry(e,john)$, where $Cry$ becomes a 2-place predicate taking an event variable and its subject as arguments.

Later on, based on Davidson's theory of events, Parsons proposed the Neo-Davidsonian event semantics in \cite{parsons91}. In Parsons' proposal, several modifications were made. First, event participants were added in more detail via thematic roles; second, besides action verbs, state verbs were also associated with an abstract variable; furthermore, the concepts of holding and culmination for events, the decomposition of events into subevents, and the modification of subevents by adverbs were investigated.

%\subsection{Evidence for Underlying Event}

As explained in \cite{parsons91}, there are various reasons for considering event as an implicit argument in the semantic representation. Three of the most prominent ones being adverbial modifiers, perception verbs and explicit references to events.

Adverbial modifiers of a natural language sentence usually bear certain logical relations with each other. Two known properties are $permutation$ and $drop$. Take Sentence \ref{jbttsiwa} as an example.
\ex. \label{jbttsiwa} John buttered the toast slowly, deliberately, in the bathroom, with a knife, at midnight.

Permutation means the truth conditions of the sentence will not change if the order of modifiers are alternated, regardless of certain syntactic constraints; drop means if some modifiers are eliminated from the original context, the new sentence should always be logically entailed by the original one. In Parsons' theory, the above sentence is interpreted as $\exists e.(Butter(e) \wedge Subject(e,john) \wedge Object(e,toast) \wedge Slow(e) \wedge Deliberate(e) \wedge In(e,bathroom)...)$. A similar treatment for adjectival modifiers can also be found in the literature. Compared with several other semantic proposals, such as increasing the arity of verbs or higher order logic solutions, event is superior.

Aside from modifiers, perception verbs form another piece of evidence for applying event in semantic representations. As their name suggests, perception verbs are verbs that express certain perceptual aspects, such as $see$, $hear$, $feel$ , and etc. The semantics of sentences that contain perception verbs are quite different from those whose sub-clauses are built with $that$ construction. For instance, we can interpret $see$ in three different ways\footnote{``$e$'' and ``$t$'' are the same as in traditional Montague Semantics, while ``$v$'' stands for a new type for event.}:
\begin{enumerate}
\item sb. \textbf{see} sb./sth.: $e \to e \to t$, e.g., \textit{Mary sees John.}
\item sb. \textbf{see} some fact: $e \to t \to t$, e.g., \textit{Mary sees that John flies.}
\item sb. \textbf{see} some event: $e \to v \to t$, e.g., \textit{Mary sees John fly.}
\end{enumerate}
As the example shows, the first $see$ just means somebody sees somebody or something. The second $see$ indicates that Mary sees a fact, the fact is \textit{John flies}. Even if Mary sees it from TV or newspaper, the sentence is still valid. The third sentence, in contrast, is true only if Mary directly perceives the event of \textit{John flying} with her own sight.

Furthermore, natural language discourses contain examples of various forms of explicit references (mostly the $it$ anaphor) to events, for example, \textit{John sang on his balcony at midnight. \textbf{It} was horrible}.

\section{Dynamic Semantics \& Discourse Relation}

\subsection{Dynamic Semantics}
%Montague semantics $\to$ DRT $\to$ Philippe's proposal
In the 1970s, based on the principle of compositionality, Richard Montague combined First Order Logic, $\lambda$-calculus, and type theory into the first formal natural language semantic system, which could compositionally generate semantic representations. This framework was formalized in \cite{montague70english}, \cite{montague70ug}, and \cite{montague73}. By convention, it is named Montague Grammar (MG).

However, despite its huge influence in semantic theory, MG was designed to handle single sentence semantics. Later on, some linguistic phenomena, such as anaphora, donkey sentences, and presupposition projection began to draw people's attention from MG to other approaches, such as dynamic semantics, which has a finer-grained conception of meaning. By way of illustration, we can look at the following ``donkey sentence'', which MG fails to explain:

\ex.  \a. A farmer$_{1}$ owns a donkey$_{2}$. He$_{1}$ beats it$_{2}$.
	\b.* Every farmer$_{1}$ owns a donkey$_{2}$. He$_{?}$ beats it$_{?}$.

In the traditional MG, the meaning of a sentence is represented as its truth conditions, that is the circumstances in which the sentence is true. However, in dynamic semantics, the meaning of a sentence is its context change potential. In other words, meaning is not a static concept any more, it is viewed as a function that always builds new information states out of the old ones by updating the current sentence. Some of the representative works, which emerged since the 1980s, include File Change Semantics \cite{FCS}, Discourse Representation Theory (DRT) \cite{DRT}, and Dynamic Predicate Logic (DPL) \cite{DPL}.

\subsection{A New Approach to Dynamics}\label{philippe}

Recently in \cite{philippe}, de Groote introduced a new framework, which integrates a notion of context into MG, based only on Church's simply-typed $\lambda$-calculus. Thus the concept of discourse dynamicscan be embedded in traditional MG without any other specific definitions as is the case in other dynamic systems.

In DRT, the problem of extending quantifier scope is tackled by introducing sets of reference markers. These reference markers act both as existential quantifiers and free variables. Because of their special status, variable renaming is very important when combining DRT with MG. The framework in \cite{philippe} is superior in the computational aspect because the variable renaming has already been solved with the simply typed $\lambda$-calculus. Further more, every new sentence is only processed under the environment of the previous context in DRT, but \cite{philippe} proposed to evaluate a sentence based on both left and right contexts, which would be abstracted over its meaning.

In Church's simple type theory, there are only two atomic types: ``$\iota$'', denoting the type of individual; ``$o$'', denoting the type of proposition\footnote{Here we follow the original denotation in \cite{philippe}, but actually there is no great difference between ``$\iota$'', ``$o$'' (Church's denotation) and ``$e$'', ``$t$'' (Montague's denotation).}. The new approach adds one more atomic type ``$\gamma$'', to express the left contexts, thus the notion of dynamic context is realized. Consequently, as the right context could be interpreted as a proposition given its left context, its type should be $\gamma \rightarrow o$. For the same reason, the whole discourse could be interpreted as a proposition given both its left and right contexts. Assuming $s$ and $t$ is respectively the syntactic category for sentence and discourse, their semantic interpretations are:
\[\llbracket s \rrbracket = \gamma \rightarrow (\gamma \rightarrow o) \rightarrow o, 
\llbracket t \rrbracket = \gamma \rightarrow (\gamma \rightarrow o) \rightarrow o\]
In order to conjoin the meanings of sentences to obtain the composed meaning of a discourse, the following formula is proposed:
\begin{equation}
\llbracket D.S \rrbracket = \lambda e \phi . \llbracket D \rrbracket e (\lambda e'. \llbracket S \rrbracket e' \phi)
\label{com_original}
\end{equation}
in which $D$ is the preceding discourse and $S$ is the sentence currently being processed. The updated context $D.S$ also possesses the same semantic type as $D$ and $S$, it has the potential to update the context. Turning to DRT, if we assume ``$x_1, x_2, \cdots$'' are reference markers, and ``$C_1, C_2, \cdots$'' are conditions, the corresponding $\lambda$-term for a general DRS in the new framework should be:
\[\lambda e \phi.\exists x_1 \cdots x_n.C_1 \wedge \cdots C_m \wedge \phi e' \footnote{Here, ``$e'$'' is a left context made of ``$e$'' and the variables ``$x_1, x_2, x_3\cdots$''. Its construction depends on the specific structure of the context, for more details see \cite{philippe}.}\]
To solve the problem of anaphoric reference, \cite{philippe} introduced a special choice operator. The choice operator is represented by some oracles, such as $sel_{he}, sel_{she},...$. It takes the left context as argument and returns a resolved individual. In order to update the context, another operator ``::'' is introduced, which adds new accessible variables to the processed discourse. For instance, term ``$a::e$'' actually is interpreted as ``$\left\lbrace a \right\rbrace \bigcup e$'' mathematically. In other words, we can view the list as the discourse referents in DRT.

Finally, let us look at a compositional treatment of Discourse \ref{jkmphilippe} according to the above formalism. The detailed type and representation for each lexical entry is presented in the following table:%Table \ref{toygrammar}.

%Finally, there is an example of compositional treatment to Discourse \ref{jkmphilippe} with the above formalism, the detailed type and representation for each lexical entry could be found in the following table and derivation:%Table \ref{toygrammar}.

%\begin{table}[ht]
\begin{center}
\begin{tabular}{|c|c|c|}
\hline
\textbf{Word} & \textbf{Type} & \textbf{Semantic Interpretation}\\ \hline
John/Mary & \multirow{2}{*}{$(\iota \rightarrow \llbracket s \rrbracket)\rightarrow \llbracket s \rrbracket$} & $\lambda \psi e \phi. \psi \textbf{j}/\textbf{m} e(\lambda e.\phi (\textbf{j}/\textbf{m}::e))$\\ \cline{1-1} \cline{3-3}
she & & $\lambda \psi e \phi. \psi (sel_{she}e) e \phi$\\ \hline
kisses & $\llbracket np \rrbracket \rightarrow \llbracket np \rrbracket \rightarrow \llbracket s \rrbracket$ & $\lambda o s. s(\lambda x. o(\lambda y e \phi . Kiss(x,y)\wedge \phi e))$\\ \hline
smiles & $\llbracket np \rrbracket \rightarrow \llbracket s \rrbracket$ & $\lambda s. s(\lambda x e \phi. Smile(x) \wedge \phi e)$\\ \hline
\end{tabular}
\end{center}
%\caption{Toy Grammar for Dynamics}
%\label{toygrammar}
%%\vspace{-5ex}
%\end{table}

\ex. \label{jkmphilippe} John kisses Mary. She smiles.

$(\llbracket kisses \rrbracket \llbracket Mary \rrbracket) \llbracket John \rrbracket \Rightarrow _ \beta \lambda e \phi. Kiss(j, m) \wedge \phi(m::j::e)$\\
$\llbracket smile \rrbracket \llbracket she \rrbracket \Rightarrow _ \beta \lambda e \phi. Smile(sel_{she}(e)) \wedge \phi (e)$\\
$\llbracket D.S \rrbracket \Rightarrow _ \beta \lambda e \phi.\textbf(Kiss(j,m) \wedge Smile(sel_{she}(j::m::e)) \wedge \phi(j::m::e))$

\subsection{Discourse Relations \& Discourse Structure}\label{drds}

Since the emergence of dynamic semantics, people have been changing their opinion on the notion of meaning. Based on that, many researchers working on multiple-sentence semantics have studied an abstract and general concept: discourse structure, in other words, the rhetorical relations, or coherence relations (\cite{hobbs1985}, \cite{mann1988rhetorical}, \cite{asher1993reference}). Representative theories include Rhetorical Structure Theory (RST) and Segmented Discourse Representation Theory (SDRT). The idea that an internal structure exists in discourse comes naturally. Intuitively, in order for a context to appear natural, its constituent sentences should bear a certain coherence with each other, namely discourse relations (DRs). That is also why it is not the case that any two random sentences can form a natural context.

It is still an open question to identify all existing DRs. But it is generally agreed there are two classes of DRs, namely the coordinating relations and the subordinating relations. The former includes relations like Narration, Background, Result, Parallel, Contrast, etc., while relations such as Elaboration, Topic, Explanation, and Precondition belong to the latter type. The distinction of two classes of DRs also has intuitive reasons.

For instance, the function of a sentence over its context could be to introduce a new topic or to support and explain a topic. Thus the former plays a subordinate role, and the latter plays a coordinate role together with those that function in the similar way (supporting or explaining). In addition, it is a even more complicated task to determine which DRs belong to which class. \cite{asher2005subordinating} provides some linguistic tests to handle this problem and analyzes some deeper distinctions between these two classes.
%For instance, the function of some sentences in the context is to pose a new topic, while the function of others is to support and explain that topic. Thus the previous ones play a subordinate role to the rest, while the others play a coordinate role among each other. On the other hand, it is a even more complicated task to determine which DRs belong to which class. \cite{asher2005subordinating} provides some linguistic tests to handle this problem and analyzes some deeper distinctions between these two classes.

\begin{figure}%[htbp]
\begin{center}
\includegraphics [totalheight=.23\textheight] {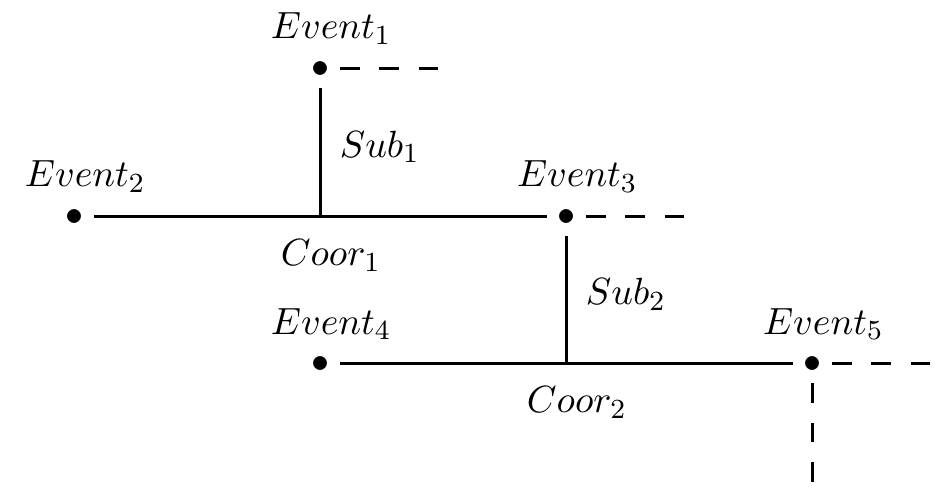}
%\Tree [.Node\_1 Node\_2 [.Node\_3 Node\_4 [.Node\_5 {\ldots} ] ] ]
%%\Tree [.Node\_1 Node\_2 [.Node\_3 Node\_4 [.Node\_5 [\qroof{a simple tree}. ] ] ] ]
\end{center}
\caption{Graph Structure Example}
\label{Graph_Structure_Example}
%\vspace{-4ex}
\end{figure}

The reason that we introduced different types of DRs is because we can construct a more specific discourse hierarchy based on it. The hierarchy can aid in the resolution of some semantic or pragmatic phenomena like anaphora. The original theoretical foundation of this idea dates back to \cite{polanyi1988formal}, which says that in a discourse hierarchy, only constituents at \textit{accessible nodes} can be integrated into the updated discourse structure. By convention, a subordinating DR creates a vertical edge and coordinating DR a horizontal edge. The accessible nodes are all located on the right frontier in the hierarchy. This is also known as the \textit{Right Frontier Constraint}. For instance, in Figure \ref{Graph_Structure_Example},
$Event_1$, $Event_3$ and $Event_5$ are on the right frontier, so they stay accessible for further attachments. However, $Event_2$ and $Event_4$ are blocked, which indicates that variables in these two nodes cannot be referenced by future anaphora.

So far, we are clear about the fact that discourses do have structures. By comparing with other dynamic semantic treatments of phenomena such as pronouns and tense, we can identify advantages of using DRs. For further illustration, we use the example from \cite{lascarides:asher:1993}:
\ex.\label{original.dis} \a. John had a great evening last night.
	\b. He had a great meal.
	\c.\label{c} He ate salmon.
	\d.\label{d} He devoured lots of cheese.
	\e. He won a dancing competition.
	\f.\label{f} *It was a beautiful pink.
\vspace{-1ex}

\begin{figure}%[htbp]
\begin{center}
\includegraphics [totalheight=.23\textheight] {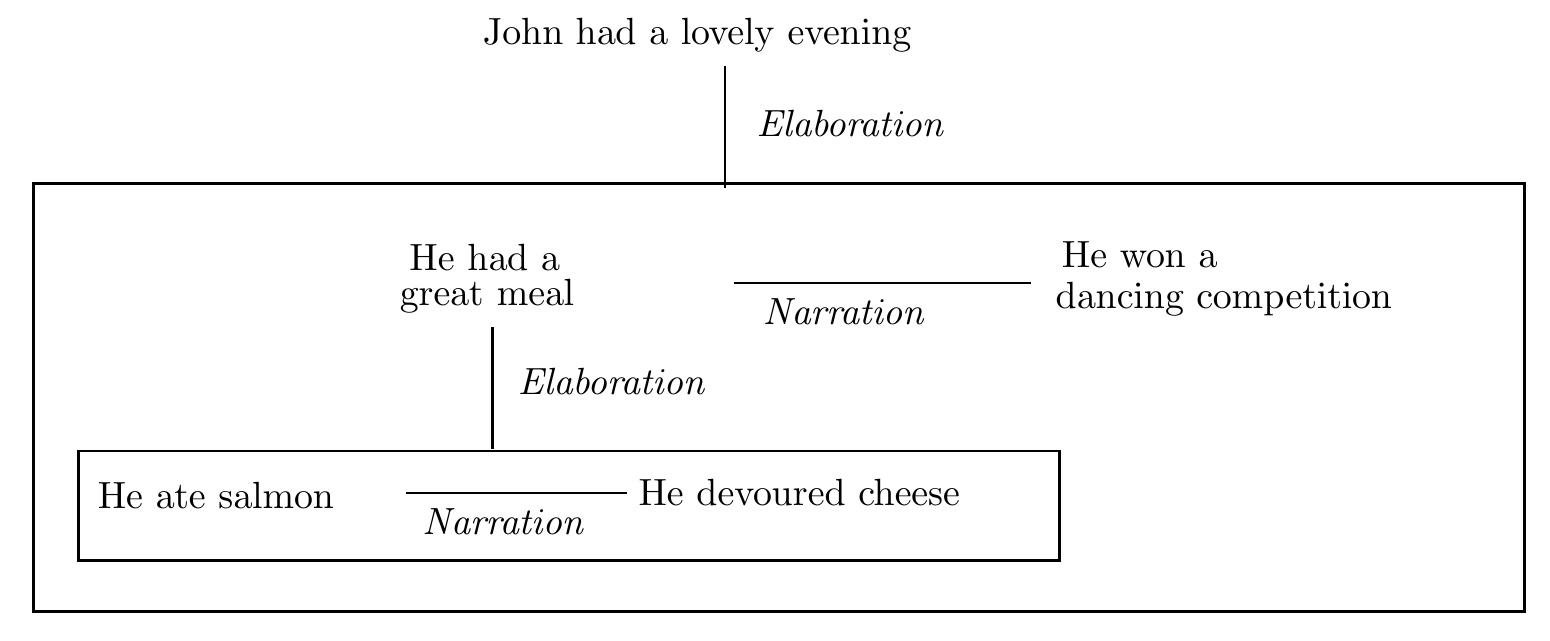}
\caption{Discourse Hierarchy of \ref{original.dis}}
\label{discourse_hierarchy}
\end{center}
%\vspace{-5ex}
\end{figure}

Traditional dynamic semantic frameworks, such as DRT, will totally accept Discourse \ref{original.dis}, because there is no universal quantification or negation to block any variable. The pronoun \textit{it} in \ref{f} can either refer back to \textit{meal}, or \textit{salmon}, or \textit{cheese}, or \textit{competition}. Normally \textit{pink} will only be used to describe \textit{salmon}, which is in the candidate list. However, sentence \ref{f} does sound unnatural to English-speaking readers. Here discourse structure can help to explain. If we construct the discourse hierarchy according to different types of DRs introduced above, we obtain the graph in Figure \ref{discourse_hierarchy}.

Thus, it is clear that \ref{f} is not able to be attached to \ref{c}, where \textit{salmon} is located. Relation between \ref{c} and \ref{d} is a Narration, which is of coordinating type, so \ref{c} is blocked for further reference. In addition, many linguistic phenomena other than anaphora can be better explained with discourse structure and the right frontier constraint, such as presupposition projection, definite descriptions, and word sense disambiguation.

%Right frontier constraints

%Two types of discourse relations: subordinating, coordinating

\section{Event in Dynamic Semantics}

So far, we have first presented the advantages of using events in semantic analysis over traditional MG, then the motivation for dynamics in discourse semantics, and finally the need for DRs in more subtle semantic processing. %However, currently there is no specific formalism which well combines events, dynamics and discourse structure.
In this section, we propose a framework that compositionally constructs event-style discourse semantics, with various DRs and the accessibility constraint (right frontier constraint) embedded. First, we explain how to build representations of single sentences. After that, the meaning construction for discourse, which is based on its component sentences, will be presented.

\subsection{Event-based Sentential Semantics}

As we showed in Section \ref{event_intro}, the implicit event argument helps to handle many linguistic phenomena, such as adverbial modifications, sentential anaphora (\textit{it}) resolution. With the notion of thematic relations, the verb predicate will take only one event variable as argument, instead of multiple variables, each representing a thematic role relation. Most of the current theories only describe event semantics from a philosophical or pure linguistic point of view, and the corresponding semantic representations are provided without concrete computational constructions. That is what our proposal focuses on. Before we introduce our framework, some assumptions need to be specified.

Thematic roles have been used formally in literature since \cite{gruber1965studies}. However, to determine how many thematic roles are necessary is still an open question. In addition, indicating exactly which part of a sentence correlates to which thematic role is also a difficult task. In our framework, we only consider the most elementary and the most widely accepted set of thematic roles. The roles and their corresponding syntactic categories are listed in the following table:%Table \ref{thematic_roles}.
%\begin{table}[thdp]
\begin{center}
\begin{tabular}{|c|c|}
\hline
\textbf{Thematic Role} & \textbf{Syntactic Correspondence}\\ \hline
Agent & Subject\\ \hline
Theme & Direct object; subject of ``$be$''\\ \hline
Goal & Indirect object, or with ``$to$''\\ \hline
Benefactive & Indirect object, or with ``$for$''\\ \hline
Instrument & Object of ``$with$''; subject\\ \hline
Experiencer & Subject\\ \hline
Location/Time & With ``$in$'' or ``$at$''\\ \hline
\end{tabular}
\end{center}
%\caption{Basic Thematic Roles}
%\label{thematic_roles}
%%\vspace{-5ex}
%\end{table}

In \cite{parsons91}, the author provides a template-based solution to construct semantic representations with events. Sentences are first classified into different cases based on their linguistic properties, such as $passive$, $perceptive$, $causative$, $inchoative$, etc. Then a unique template is assigned to each case; the number, types and positions of arguments are specifically designed for that template. In our proposal, we will also use templates,  but a much simpler version. The templates only contain the most basic thematic roles for certain verbs. They are subject to modification and enhancement for more complicated cases. For instance, the template for the verb \textit{smile} only contains one agent role, while the verb \textit{kiss} contains both agent and theme roles.

Furthermore, our proposal will generalize the ontology of event variables. So to speak, at the current stage we make no distinction between $events$, $states$ and $processes$, for the sake of simplicity (their linguistic differences are described in \cite{parsons91}). So there will just be one unique variable, representing the underlying event, or state, or process, indicated by the verb. There is a simple example:
\ex.\label{jkm} John kisses Mary in the plaza.

Under event semantics, the semantic representation for Sentence \ref{jkm} should be:
\[\exists e.(Kiss(e) \wedge Ag(e, john) \wedge Pat(e, mary) \wedge Loc(e, plaza))\footnote{$Ag$ stands for Agent, $Pat$ for Patient and $Loc$ for Location}.\]
In order to obtain the above representation compositionally, we use the following semantic entries for words in the lexicons:
\begin{center}
$\llbracket John \rrbracket = john$$\quad\quad\quad$$\llbracket Mary \rrbracket = mary$\\
$\llbracket kiss \rrbracket = \lambda o s e.(Kiss(e) \wedge Ag(e, s) \wedge Pat(e, o))$\\
$\llbracket in\_the\_plaza \rrbracket = \lambda P e.(P(e) \wedge Loc(e, plaza) )$\footnote{It is of course possible to break down the interpretation construction of ``\textit{in the plaza}'' into a more detailed level by providing entries for each word, but we give the compound one for the whole PP  just for simplification.}
\end{center}
Thus, by applying the above four entries to one another in a certain order\footnote{The function-argument application can be obtained via shallow syntactic processing.}, we can compute the semantic representation of \ref{jkm} step by step:
\begin{enumerate}
\item $\llbracket kiss \rrbracket \llbracket Mary \rrbracket\\
\Rightarrow _ \beta \lambda s e.(Kiss(e) \wedge Ag(e, s) \wedge Pat(e, mary))$
\item $(\llbracket kiss \rrbracket \llbracket Mary \rrbracket)\llbracket John \rrbracket\\
\Rightarrow _ \beta \lambda e.(Kiss(e) \wedge Ag(e, john) \wedge Pat(e, mary))$
\item $\llbracket in\_the\_plaza \rrbracket((\llbracket kiss \rrbracket \llbracket Mary \rrbracket)\llbracket John \rrbracket)\\
\Rightarrow _ \beta \lambda P e.(P(e) \wedge Loc(e, plaza) )(\lambda e.(Kiss(e) \wedge Ag(e, john)\wedge Pat(e, mary)))$\\
$\Rightarrow _ \beta \lambda e'.(\lambda e.(Kiss(e) \wedge Ag(e, john) \wedge Pat(e, mary)))(e') \wedge Loc(e, plaza)$\\
$\Rightarrow _ \beta \lambda e.(Kiss(e) \wedge Ag(e, john) \wedge Pat(e, mary) \wedge Loc(e, plaza))$
\end{enumerate}
At this point, the event variable ``$e$'' is not yet instantiated as an existential quantifier. To realize that, we can simply design an EOS (End Of Sentence) operator\footnote{This could be a comma, full stop, exclamation point, or any other punctuation marks.}, to which the partial sentence representation could be applied:
\begin{center}
$\llbracket EOS \rrbracket = \lambda P.\exists e.P(e)$\\
\end{center}
As a result, the last step would be:
\begin{enumerate}
\setcounter{enumi}{3}
\item $\llbracket EOS \rrbracket(\llbracket in\_the\_plaza \rrbracket((\llbracket kiss \rrbracket \llbracket Mary \rrbracket)\llbracket John \rrbracket))$\\
$\Rightarrow _ \beta \lambda P.\exists e.P(e)(\lambda e.(Kiss(e) \wedge Ag(e, john) \wedge Pat(e, mary) \wedge Loc(e, plaza)))$\\
$\Rightarrow _ \beta \exists e. (Kiss(e) \wedge Ag(e, john) \wedge Pat(e, mary) \wedge Loc(e, plaza))$
\end{enumerate}

In the above solution, the adverbial modifier \textit{in the plaza} is handled in the manner that is traditional for intersective adjectives (e.g., \textit{tall}, \textit{red}). With a similar formalism, any number of intersective adverbial modifiers can be added to the event structure, as long as the corresponding lexical entry is provided for each modifier.

The event variable, which is embedded in the verb, is the greatest difference between our framework and MG. From a computational point of view, we need to pass the underlying event variable from the verb to other modifiers. As a result, we first use the $\lambda$-operator for the event variable in the verb interpretation, then the EOS to terminate the evaluation and instantiate the event variable with an existential quantifier. Another framework which compositionally obtain an event-style semantics is \cite{Bos2009GSCL}, which introduces the existential quantifier for the event at the beginning of interpretation.%This solution also supports the idea that the event comes from the verb. In the following section, in which we encounter multiple-sentence situations, more examples will be presented.

\subsection{Event-based Discourse Semantics}
\label{event_discourse}
In the previous part, we showed how to compute single sentence semantics with events. In this section we will combine event structure with dynamic semantics, extending our formalism to discourse.

As explained in Section \ref{philippe}, \cite{philippe} expresses dynamics in MG by introducing the concept of left and right contexts. We adopt the idea, inserting the left and right contexts into our semantic representations. Thus we bestow upon our event-based formalism the potential to be updated as in other dynamic systems. To achieve this, we modify the lexical entries in the previous section as following:
\begin{center}
$\llbracket John \rrbracket = john$$\quad\quad\quad$$\llbracket Mary \rrbracket = mary$\\
$\llbracket kiss \rrbracket = \lambda o s e a b.(Kiss(e) \wedge Ag(e, s) \wedge Pat(e, o) \wedge b (e::a))$\\
$\llbracket in\_the\_plaza \rrbracket = \lambda P e a b.(P e a b \wedge Loc(e, plaza) )$
\end{center}
Here, in contrast to the notation used in \cite{philippe}, ``$a$'' stands for the left context and ``$b$'' stands for the right context. In our logical typing system, we use type ``$v$'' for the event variable, and type ``$\alpha$'' for the left context. Types ``$e$'' and ``$t$'' have the same meaning as convention. In an additional departure from the formalism in \cite{philippe}, we assume the left context contains the accessibility information of previous event variables, instead of individual variables. That is why we keep using the original interpretations for $John$ and $Mary$, instead of inserting the constants ``$john$'' and ``$mary$'' in the left context list structure. However, the list constructor ``::'' does have a similar meaning. The only difference with the constructor in \cite{philippe} is that our ``::'' takes an event variable and the left context as arguments, while the previous one takes an individual variable and the left context. Given the lexical entries above, the semantic representation with a dynamic potential for Sentence \ref{jkm} becomes:
\begin{enumerate}
\item $\llbracket in\_the\_plaza \rrbracket((\llbracket kiss \rrbracket \llbracket Mary \rrbracket)\llbracket John \rrbracket)$\\
$\Rightarrow _ \beta \lambda e a b.(Kiss(e) \wedge Ag(e, john) \wedge Pat(e, mary) \wedge Loc(e, plaza) \wedge b (e::a))$
\end{enumerate}
Its semantic type also changes from ``$v \to t$'' into ``$v \to (\alpha \to (\alpha \to t) \to t)$''\footnote{``$\alpha \to t$'' is the type for the right context, represented by variable ``$b$''.}. In order to terminate the semantic processing, we need a new EOS symbol:
\begin{center}
$\llbracket EOS \rrbracket = \lambda P.\exists e.P e A B$\\
\end{center}
Thus we obtain the final interpretation:
\begin{enumerate}
\setcounter{enumi}{1}
\item $\llbracket EOS \rrbracket(\llbracket in\_the\_plaza \rrbracket((\llbracket kiss \rrbracket \llbracket Mary \rrbracket)\llbracket John \rrbracket))$\\
$\Rightarrow _ \beta \exists e. (Kiss(e) \wedge Ag(e, john) \wedge Pat(e, mary) \wedge Loc(e, plaza) \wedge B (e::A))$
\end{enumerate}
The ``$A$'' and ``$B$'' in the EOS and above formula are not variables any more. They are just constants of type ``$\alpha$'' and ``$\alpha \to t$'', respectively, which have the effect of freezing the left and right contexts. We can see that the new representation does not seem different from the previous version. That is, of course, because although we embed the dynamic potential into the entries, we are still evaluating single sentence semantics. The power of dynamics will not show up until the case becomes more complicated. So let us consider the following discourse:
\ex.\label{jkmss} \a.\label{jkmssa}John kisses Mary in the plaza.
	\b.\label{jkmssb} She smiles.

To handle Example \ref{jkmss}, we need to provide two more entries:
\begin{center}
$\llbracket she \rrbracket = \lambda P e a b. P(Sel (a)) e a b$\\
$\llbracket smile \rrbracket = \lambda s e a b. (Smile(e) \wedge Ag(e, s) \wedge b (e::a))$\\
\end{center}

Inspired by \cite{philippe}, the interpretation of $she$ is made by an external function: $Sel$. 
%But the function in our framework is supposed to first select the correct event variable, then the corresponding individual variable. 
This function is supposed to work over a structured representation of the discourse: we claim that individual variables are defined in the scope of event variables. Thus the resolution of this anaphora must be first do by
picking out an event variable, and, through this event, choose the correct individual variable following the previous.

We also apply a type-raising representation for NP ($she$), because we need to pass the selection function $Sel$ for further processing. Similar type-raising version of $John$ and $Mary$ could also be constructed. After type raising, the only thing that needs to be changed is the order of argument application, and the resulting logic term will exactly be the same.

So, getting back to Discourse \ref{jkmss}, we can first obtain the representations for \ref{jkmssa} and \ref{jkmssb} independently:
\begin{enumerate}
\item $\llbracket in\_the\_plaza \rrbracket((\llbracket kiss \rrbracket \llbracket Mary \rrbracket)\llbracket John \rrbracket)$\\
$\Rightarrow _ \beta \lambda e a b.(Kiss(e) \wedge Ag(e, john) \wedge Pat(e, mary) \wedge Loc(e, plaza) \wedge b (e::a))$
\item $\llbracket she \rrbracket \llbracket smile \rrbracket$\\
$\Rightarrow _ \beta \lambda e a b .(Smile(e) \wedge Ag(e, Sel (a)) \wedge b (e::a))$
\end{enumerate}
Now the problem is how to combine the two interpretations to yield the discourse semantics. \cite{philippe} uses Formula \ref{com_original} to merge sentence interpretations, which takes the previous discourse and the current sentence as input, returns a new piece of updated discourse. Same as DRT, there is no rhetorical relation involved in \cite{philippe}. However, this paper goes one step further, aiming to encode the discourse structure and event accessibility relations between different sentences.

Hence, we make another assumption here: $discourse$ and $sentence$ are distinct semantic entities, they have different types and meaning evaluations. Every discourse contains certain rhetoric relations, while single sentences should be able to be interpreted without those relations. That is because we consider the discourse structure as a production from sentence composition. Unlike in Formula \ref{com_original}, where discourse $D$ and sentence $S$ have exactly the same semantic properties, we assign them different types. Drawing from the above assumption, the event variables should be instantiated into existential quantifier in discourse only, while they are still of $\lambda$-forms in sentences. By way of illustration, the followings are the most general representations for sentence and discourse:
{\small \begin{center}
$\llbracket S \rrbracket = \lambda e a b.(Pred(e) \wedge ... \wedge b a)$\\
$\llbracket D \rrbracket = \lambda a b.\exists e_{1} e_{2}...(Pred_{1}(e_{1}) \wedge Pred_{2}(e_{2}) \wedge ... \wedge Rel_{1}(e_{i}, e_{j}) \wedge Rel_{2}(e_{m}, e_{n}) \wedge... \wedge b a')$\footnote{The left context ``$a'$'' in the representation is a complicated structure containing the event accessibility relation. There will be further examples showing how to create ``$a'$'' from ``$a$'' and other event variables.}
\end{center}}
Please note that the interpretation for discourse does not only contain ``$a'$'', where accessibility relations are located; but also various rhetoric relations, represented by $Rel_{1}$, $Rel_{2}$ and so on. Those rhetoric relations, as we discussed in Section \ref{drds}, can be classified into either subordinating or coordinating. They have completely different effects in shaping the discourse structure graphs, which determines the accessibility relations. Here we do not care about how many different discourse relations there are (such as Narration, Background, Elaboration, etc.), we just assume if there is a relation, it must belong to one of the two classes. And those rhetoric relations are added only during the meaning merging process. As a consequence, we propose two sets of composition functions, according to different types of DRs.

\subsubsection{Subordinating Composition Functions}

\begin{figure}%[htbp]
\begin{center}
\includegraphics [totalheight=.23\textheight] {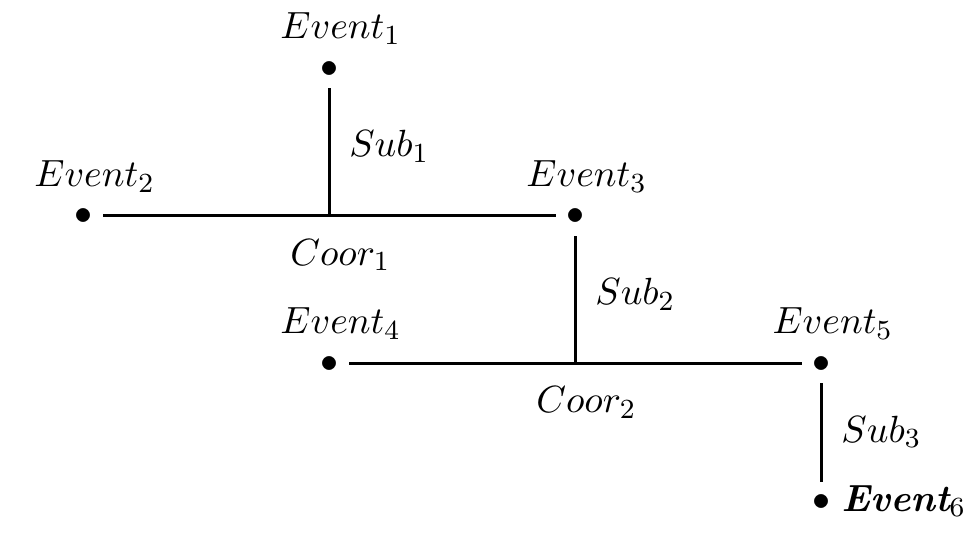}
%\Tree [.Event\_1 Event\_2 [.Event\_3 Event\_4 [.Event\_5 \textbf{Event\_6} ] ] ]
%%\Tree [.Node\_1 Node\_2 [.Node\_3 Node\_4 [.Node\_5 [\qroof{a simple tree}. ] ] ] ]
\caption{Graph Structure with Subordinating Relations}
\label{Sub_Graph_Structure2}
\end{center}
%\vspace{-5ex}
\end{figure}

Based on the right frontier constraint, for those discourses and sentences which are connected by subordinating DRs, all accessible nodes in the previous discourse remain the same, meanwhile the new sentence will be inserted as accessible in the updated discourse. For example  in Figure \ref{Sub_Graph_Structure2},
when $Event_6$ is added into the discourse by a subordinating relation with $Event_5$, the current accessible nodes include $Event_1$, $Event_3$, $Event_5$ and $Event_6$. Hence, we introduce the composition functions for subordinating DRs as follows:
\begin{equation}
\llbracket Sub_{Bas} \rrbracket = \lambda D S a b. D a (\lambda a'. \exists e.( S e a' b))
\label{com_sub_basic}
\end{equation}
\begin{equation}
\llbracket Sub_{Adv} \rrbracket = \lambda D S a b. D a (\lambda a'. \exists e. ( (S e a' b) \wedge Rel(Sel(a'), e))
%\lambda a_{1} b_{1}. D a_{1} (\lambda a_{2}. \exists e. ( (S e a_{2} b_{1}) \wedge Rel(Sel(a_{2}), e))
\label{com_sub_advanced}
\end{equation}
We suppose that every sentence needs to combine with a previous discourse to form a new discourse, also including the first sentence in the context. However, the first sentence could only be combined with an empty discourse:
\[\llbracket Empty \rrbracket = \lambda a b. b a\]
which in fact contains no context information at all, it is created just for computational reason. That's why we design two composition functions \ref{com_sub_basic} and \ref{com_sub_advanced}, namely the $Sub_{Bas}$ and the $Sub_{Adv}$, to respectively handle the first sentence case and all other situations. Now we will construct the interpretation of \ref{jkmss}, as an illustration for our composition functions. Suppose \ref{jkmssa} and \ref{jkmssb} hold a subordinating relation between each other\footnote{Here is just an assumption, our system does not account how to determine the DRs, we only focus on encoding those relations.}, then in order to obtain the whole representation for \ref{jkmss}, we first need to combine \ref{jkmssa} with the empty discourse by $Sub_{Bas}$, then combine the result with \ref{jkmssb} by $Sub_{Adv}$.

\begin{enumerate}[1.]
\item $\llbracket Sub_{Bas} \rrbracket \llbracket Empty \rrbracket \llbracket \ref{jkmssa} \rrbracket$\footnote{We omit some internal thematic structures for \ref{jkmssa} just for a clear view of the logic terms. The same omission will also be carried out for \ref{jkmssb}.}\\
$\Rightarrow _ \beta \lambda a_{1} b_{1}. (\lambda a_{3} b_{3} . b_{3} a_{3}) a_{1} (\lambda a_{2}. \exists e. (\lambda e' a_{4} b_{4}.(Kiss(e') \wedge... \wedge b_{4} (e'::a_{4})) e a_{2} b_{1}  ) )$\\
$\Rightarrow _ \beta \lambda a_{1} b_{1}. (\lambda b_{3} . b_{3} a_{1}) (\lambda a_{2}. \exists e.  (Kiss(e) \wedge... \wedge b_{1} (e::a_{2})))$\\
$\Rightarrow _ \beta \lambda a_{1} b_{1}. \exists e.  (Kiss(e) \wedge... \wedge b_{1} (e::a_{1} ))$
\end{enumerate}
This step does two things. First, it instantiates the event variable from \ref{jkmssa} into an existential quantifier. In addition, it inserts the new event argument into the accessible list of the left context. Because the empty discourse does not contain any variable in its left context, the list construction is fairly simple, we just need a naive ``push-in'' operation.
\begin{enumerate}
\setcounter{enumi}{1}
\item $\llbracket Sub_{Adv} \rrbracket (\llbracket Sub_{Bas} \rrbracket \llbracket Empty \rrbracket \llbracket \ref{jkmssa} \rrbracket) \llbracket \ref{jkmssb} \rrbracket$\\
$\Rightarrow _ \beta \lambda a_{1} b_{1}. (\lambda a_{3} b_{3}. \exists e_{1}.  (Kiss(e_{1}) \wedge... \wedge b_{3} (e_{1}::a_{3} ))  ) a_{1} (\lambda a_{2}. \exists e. ( ( ( \lambda e_{2} a_{4} b_{4} .$

$(Smile(e_{2}) \wedge ... \wedge b_{4} (e_{2}::a_{4}))  ) e a_{2} b_{1}) \wedge Rel(Sel(a_{2}), e)))$\\
$\Rightarrow _ \beta \lambda a_{1} b_{1}. (\lambda b_{3}. \exists e_{1}.  (Kiss(e_{1}) \wedge... \wedge b_{3} (e_{1}::a_{1} ))  ) (\lambda a_{2}. \exists e. ( Smile(e) \wedge ... \wedge b_{1} (e::a_{2})  \wedge Rel(Sel(a_{2}), e)))$\\
$\Rightarrow _ \beta \lambda a_{1} b_{1}. \exists e_{1}.  (Kiss(e_{1}) \wedge... \wedge \exists e. ( Smile(e) \wedge ... \wedge b_{1} (e::e_{1}::a_{1} )  \wedge Rel(Sel(e_{1}::a_{1} ), e))) $\\
$= \lambda a_{1} b_{1}. \exists e_{1} e_{2}.  (Kiss(e_{1}) \wedge... \wedge Smile(e_{2}) \wedge ... \wedge b_{1} (e_{2}::e_{1}::a_{1} )  \wedge Rel(Sel(e_{1}::a_{1} ), e_{2})) $
\end{enumerate}
Suppose the selection function $Sel$ is able to pick the correct event variable out of the accessible list, then our desired DRs and accessibility relation would be successfully encoded in the final logic formula. There are two more remarks for the subordinating composition functions: 1. no new event variable is created during the meaning composition, but all event variables with the $\lambda$-operator will be instantiated as existential quantifiers; 2. the composing process will not change the accessibility condition in the previous discourse, only a new accessible node is added.

\subsubsection{Coordinating Composition Functions}

\begin{figure}%[htbp]
\begin{center}
\includegraphics [totalheight=.19\textheight] {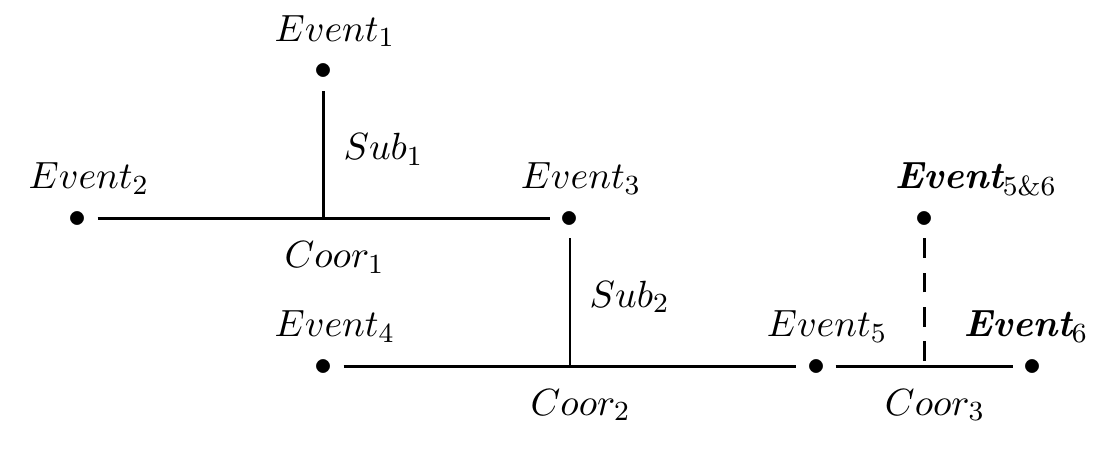}
%\Tree [.Event\_1 Event\_2 [.Event\_3 Event\_4 Event\_5 \textbf{Event\_6} ] ]
%%\Tree [.Node\_1 Node\_2 [.Node\_3 Node\_4 [.Node\_5 [\qroof{a simple tree}. ] ] ] ]
\caption{Graph Structure with Coordinating Relations}
\label{Sub_Graph_Structure}
\end{center}
%\vspace{-5ex}
\end{figure}

Again, let's first analyze the effect of coordinating DRs on accessibility structure. When a new node is added to an existing discourse with coordinating relation, a horizontal edge is built, as shown in Figure \ref{Sub_Graph_Structure}, $Event_6$ and $Event_5$ for example.
At the same time, an abstract variable node - $Event_{5\&6}$, is created. This is a distinct property compared to subordinating DRs. We need the new abstract node because in many cases the anaphora $it$ could only be resolved with reference to a set of sentences connected with coordinating relations, as in: 

\ex.\label{exe} Mary stumbled her ankle. She twisted it. John did so too.

To see more examples, see \cite{lascarides2007segmented}.

%Some real world examples are presented in \cite{lascarides2007segmented}.

Based on the above analysis, we propose the following composition functions:
{\small \begin{equation}
\llbracket Coor_{Bas} \rrbracket = \lambda D S a b. D a (\lambda a'. \exists e.( S e a' b))
\label{com_coor_basic}
\end{equation}
\begin{equation}
\llbracket Coor_{Adv} \rrbracket = \lambda D S a b. \exists e_{c}. D a (\lambda a' . \exists e. (S e (e_{c}::(Del (a'))) b ) \wedge Rel(Sel(a'), e, e_{c})  )
%\llbracket Coor\_Advanced \rrbracket = \lambda D S a_{1} b_{1}. \exists e_{c}. D a_{1} (\lambda a_{2} . \exists e. (S e (e_{c}::(Del (a_{2}))) b_{1} ) \wedge Rel(Sel(a_{2}), e, e_{c})  )
\label{com_coor_advanced}
\end{equation}}

Notice that Formula \ref{com_coor_basic} is identical to \ref{com_sub_basic} because both basic composition functions are designed only to handle the first sentence case, in which we do not really need to distinguish from different DRs (there is even no DR at all). In contrast to Formula \ref{com_sub_advanced}, the advanced subordinating function, there are three main differences in Formula \ref{com_coor_advanced}. First, apart from instantiating the event variable of current sentence, another abstract event variable ``$e_{c}$'' is created. It is directly inserted into the accessible list because the new abstract node will always be at the right frontier in the updated discourse structure. Moreover, we introduce a new function $Del$, which takes the current accessible list as argument, and deletes those nodes which will no longer be accessible in the new discourse. It works in a similar way as the $Sel$ function. Finally, the $Rel$ function takes three arguments, including the abstract variable. By doing this we can keep track of the relation between abstract variables and their component nodes.

Now let us use Discourse \ref{jkmss} again as an illustration. This time we assume the rhetoric relation between \ref{jkmssa} and \ref{jkmssb} is of a coordinating kind. Thus we will build its semantic representation with \ref{com_coor_basic} and \ref{com_coor_advanced}.
\begin{enumerate}[1.]
\item $\llbracket Coor_{Bas} \rrbracket \llbracket Empty \rrbracket \llbracket \ref{jkmssa} \rrbracket = \llbracket Sub_{Bas} \rrbracket \llbracket Empty \rrbracket \llbracket \ref{jkmssa} \rrbracket$\\
$\Rightarrow _ \beta \lambda a_{1} b_{1}. \exists e.  (Kiss(e) \wedge... \wedge b_{1} (e::a_{1} ))$
\item $\llbracket Coor_{Adv} \rrbracket (\llbracket Coor_{Bas} \rrbracket \llbracket Empty \rrbracket \llbracket \ref{jkmssa} \rrbracket) \llbracket \ref{jkmssb} \rrbracket$\\
$\Rightarrow _ \beta \lambda S a_{1} b_{1}. \exists e_{c}. (\lambda a_{3} b_{3}. \exists e_{1}.  (Kiss(e_{1}) \wedge... \wedge b_{3} (e_{1}::a_{3} ))) a_{1} (\lambda a_{2} . \exists e.((\lambda e_{2} a_{4} b_{4} .$

$(Smile(e_{2}) \wedge ... \wedge b_{4} (e_{2}::a_{4}))  )  e (e_{c}::(Del (a_{2}))) b_{1} ) \wedge Rel(Sel(a_{2}), e, e_{c}) )$\\
$\Rightarrow _ \beta \lambda S a_{1} b_{1}. \exists e_{c}. (\lambda b_{3}. \exists e_{1}.  (Kiss(e_{1}) \wedge... \wedge b_{3} (e_{1}::a_{1} )))  (\lambda a_{2} . \exists e. (Smile(e) \wedge ... \wedge b_{1} (e::e_{c}::(Del (a_{2}))  )) \wedge Rel(Sel(a_{2}), e, e_{c}) )$\\
$\Rightarrow _ \beta \lambda S a_{1} b_{1}. \exists e_{c}e_{1}e_{2}.  (Kiss(e_{1}) \wedge... \wedge Smile(e_{2}) \wedge ... \wedge b_{1} (e_{2}::e_{c}::(Del (e_{1}::a_{1}))  ) \wedge Rel(Sel(e_{1}::a_{1}), e_{2}, e_{c})) $
\end{enumerate}
As we can see from the final formula, the new event variable ``$e_{2}$'' and the abstract variable ``$e_{c}$'' are added into the accessible list. $Del$ will then eliminate the inaccessible node ``$e_{1}$'', leaving only ``$e_{2}$'' and ``$e_{c}$'' on the right frontier.

To test the validity of the proposed system, we have implemented all the above calculus in the Abstract Categorial Grammar \cite{de2001towards}.

\subsection{Comparison with Other Related Works}
Recently there are some other semantic frameworks based on discourse structure, DRT and other dynamic concepts. For example in \cite{asher2010pogodalla}, the authors expressed SDRT in a non-representational way with dynamic logic. Similar to the formalism presented in our paper, they also use the continuation calculus from \cite{philippe}, where the concepts of left and right contexts are involved for introducing dynamics. However, there are some distinctions between our work and theirs.

First of all, we use an event-style semantics for meaning representation. Consequently, the basic construct of rhetorical relation in our framework is event, in contrast with the discourse constituent unit (DCU) in \cite{asher2010pogodalla}. Event-based theory, as an independent branch of formal semantics, has been studied since a long time ago. Many lexical properties (mainly for verbs), such as tense and aspect, causative and inchoative, etc., have already been investigated in detail. By using event here, we can borrow many off-the-shelf results directly. Also, the DCUs, which are notated by $\pi$ in other discourse literatures, are not as concretely defined as events. There are cases where a single DCU contains multiple events. For instance, ``\textit{John says he loves Mary. Mary does not believe it.}''. Only with DCU, the resolution for $it$ in the second sentence will cause ambiguity. 
%First of all we use an event-style semantics and relatio in \cite{asher2010pogodalla}, DRs occur among discourse constituents in the context, which are represented by a set of labels $\pi_{1}$, $\pi_{2}$,...,$\pi_{n}$. In contrast, we use event semantics and relations between various events to encode the discourse structures. We make this choice because discourse constituent is a more obscure concept than event. Normally, each constituent strictly corresponds to a single sentence, this might cause problem. For instance in \ref{jshlm},
%\ex.\label{jshlm} \a.\label{jshlma}John says he loves Mary.
%	\b.\label{jshlmb} Mary does not believe it.

%the anaphora \textit{it} in \ref{jshlmb} could either be linked with the whole sentence of \ref{jshlma}, or the sub-part \textit{he loves Mary} of \ref{jshlma}. The event structure can properly provide two event variables to resolve the ambiguity. On the other hand, as introduced in Section \ref{event_intro}, event semantics has been studied since a long time ago. By using event structure in the framework, we can directly borrow many nice and mature results without much modification.

In addition, we and \cite{asher2010pogodalla} make different assumptions over \textit{discourse} and \textit{sentence}. The same way as in \cite{philippe}, \cite{asher2010pogodalla} views the discourse and sentence as identical semantic construct. However, as explained in Section \ref{event_discourse}, we do distinguish them as different objects. When encountering a single sentence, we should interpret it independently, without considering any discourse structure. While discourse is not simply a naive composition of component sentences. It should be their physical merging with various DRs added.

%Finally, the DRs come from different sources in the two works. \cite{asher2010pogodalla} uses key words as a DR indicator. For example in \ref{amwi},
%\ex. \label{amwi} \a. \label{amwia} A man walked in.
%	\b. \label{amwib} Then he coughed.
Finally, the DRs originate from different sources in the two works. \cite{asher2010pogodalla} uses key words as DR indicator. For example, in discourse ``\textit{A man walked in, then he coughed.}'', 	
%\cite{asher2010pogodalla} supposes the word $then$ indicates that the relation between \ref{amwia} and \ref{amwib} is $Narration$. As a result, they embed the $Narration$ relation in the interpretation of $then$. However, we believe that the DRs only reveal when sentence and discourse are combined, they should not be encoded in single sentence interpretations. Because of this, DRs are presented in the composition functions in our approach, as shown in Formula \ref{com_sub_basic}, \ref{com_sub_advanced}, \ref{com_coor_basic} and \ref{com_coor_advanced}.
\cite{asher2010pogodalla} embeds the $Narration$ relation in the interpretation of $then$. However, we believe that the DRs only be revealed when sentence and discourse are combined, they should not emerge in sentence interpretations. So DRs are presented in the composition functions (Formula \ref{com_sub_basic}, \ref{com_sub_advanced}, \ref{com_coor_basic} and \ref{com_coor_advanced}) in our work.

\section{Conclusion and Future Work}

In this paper, we have represented the accessibility relations of natural language discourse within event semantics. This approach does not depend on any specific logic, all formulas are in the traditional MG style.

We decide to use event-based structure because it is able to handle sentential anaphora resolution (e.g., $it$), adverbial modifiers and other semantic phenomena. Also, applying dynamics to event semantics may largely extend its power, which was originally developed to treat single sentences. As we know, the accessibility among sentences in discourse depends on various types of DRs. However, these DRs are usually hard to determine. We assume all DRs be classified into two types: subordinating and coordinating. Also we obtain the accessibility relation with the right frontier constraint. Based on that, we encode these DRs and the updating potential for single sentences in a First Order Logic system.

In our approach, we differentiate discourse and sentence as two distinct semantic objects. The DRs are only added during the updating process, which is realized through the set of composition functions. This choice not only has computational, but also philosophical evidences.

In this paper, we only focus on representing the DRs and accessibility in logical forms, but how to determine these DRs, or whether the DRs have a more complicated effect than the right frontier constraint could be the subjects of future works. Further more, since we tried to construct the event structure compositionally, the scoping interaction among the new quantifiers (e.g., $\exists e_{1}e_{2}...$) and previous existing ones (e.g., $\exists x_{1}x_{2}...$) also needs further investigation.

\bibliographystyle{splncs03.bst}
\bibliography{reference}

\begin{thebibliography}{10}
\providecommand{\url}[1]{\texttt{#1}}
\providecommand{\urlprefix}{URL }

\bibitem{asher1993reference}
Asher, N.: {Reference to abstract objects in discourse}. Springer (1993)

\bibitem{asher2010pogodalla}
Asher, N., Pogodalla, S.: Sdrt and continuation semantics. Proceedings of
  LENLS, Tokyo, Japan  VII (2010)

\bibitem{asher2005subordinating}
Asher, N., Vieu, L.: {Subordinating and coordinating discourse relations}.
  Lingua  115(4),  591--610 (2005)

\bibitem{Bos2009GSCL}
Bos, J.: Towards a large-scale formal semantic lexicon for text processing. In:
  Chiarcos, C., Eckart~de Castilho, R., Stede, M. (eds.) From Form to Meaning:
  Processing Texts Automatically. Proceedings of the Biennal GSCL Conference
  2009. pp. 3--14 (2009)

\bibitem{davidson67lf}
Davidson, D.: The logical form of action sentences. In: Rescher, N. (ed.) The
  Logic of Decision and Action. University of Pittsburgh Press, Pittsburgh
  (1967)

\bibitem{DPL}
Groenendijk, J., Stokhof, M.: Dynamic predicate logic. Linguistics and
  Philosophy  14(1),  39--100 (1991)

\bibitem{de2001towards}
de~Groote, P.: {Towards abstract categorial grammars}. In: Proceedings of the
  39th Annual Meeting on Association for Computational Linguistics. pp.
  252--259. Association for Computational Linguistics (2001)

\bibitem{philippe}
de~Groote, P.: Towards a montagovian account of dynamics. Proceedings of
  Semantics and Linguistic Theory XVI  (2006)

\bibitem{gruber1965studies}
Gruber, J.S.: Studies in lexical relations. Ph.D. thesis, Massachusetts
  Institute of Technology. Dept. of Modern Languages (1965)

\bibitem{FCS}
Heim, I.: File change semantics and the familiarity theory of definiteness. In:
  B\"auerle, R., Schwarze, C., von Stechow, A. (eds.) Meaning, Use, and
  Interpretation of Language, pp. 164--189. Walter de Gruyter, Berlin (1983)

\bibitem{hobbs1985}
Hobbs, J.R.: On the coherence and structure of discourse. CSLI, Center for the
  Study of Language and Information (US) (1985)

\bibitem{DRT}
Kamp, H.: A theory of truth and semantic representation. In: Groenendijk, J.,
  Janssen, T., Stokhof, M. (eds.) Formal Methods in the Study of Language, pp.
  277--322. Mathematical Centre Tracts 135, Mathematisch Centrum, Amsterdam
  (1981)

\bibitem{lascarides:asher:1993}
Lascarides, A., Asher, N.: Temporal interpretation, discourse relations and
  commonsense entailment. Linguistics and Philosophy  16(5),  437--493 (1993)

\bibitem{lascarides2007segmented}
Lascarides, A., Asher, N.: {Segmented discourse representation theory: Dynamic
  semantics with discourse structure}. Computing meaning pp. 87--124 (2007)

\bibitem{mann1988rhetorical}
Mann, W., Thompson, S.: {Rhetorical structure theory: Toward a functional
  theory of text organization}. Text-Interdisciplinary Journal for the Study of
  Discourse  8(3),  243--281 (1988)

\bibitem{montague70english}
Montague, R.: {English as A Formal Language}. Linguaggi nella societae nella
  tecnica pp. 189--224 (1970)

\bibitem{montague70ug}
Montague, R.: {Universal Grammar}. Theoria  36(3),  373--398 (1970)

\bibitem{montague73}
Montague, R.: The proper treatment of quantification in ordinary english. In:
  Hintikka, J., Moravcsik, J., Suppes, P. (eds.) Approaches to Natural
  Language. Reidel, Dordrecht (1973)

\bibitem{parsons91}
Parsons, T.: Events in the Semantics of English: A Study in Subatomic
  Semantics. MIT Press, Cambridge, MA (1991)

\bibitem{polanyi1988formal}
Polanyi, L.: A formal model of the structure of discourse. Journal of
  Pragmatics  12(5-6),  601--638 (1988)

\end{thebibliography}

\end{document}